\pdfoutput=1

\documentclass[11pt]{article}

\usepackage{ACL2023}
\usepackage[cjk]{kotex}
\usepackage{times}
\usepackage{latexsym}

\usepackage[T1]{fontenc}

\usepackage[utf8]{inputenc}

\usepackage{microtype}
\usepackage{adjustbox}
\usepackage{todonotes}
\usepackage{url}
\usepackage{graphicx}
\usepackage{booktabs}
\usepackage{amsmath}
\usepackage{enumitem}
\usepackage{algorithm}
\usepackage{algorithmic}

\usepackage{multirow}
\usepackage{multicol}
\usepackage{makecell}
\usepackage{hyperref}

\usepackage{supertabular}

\title{Towards Diverse and Effective Question-Answer Pair Generation from Children Storybooks}

\author{Sugyeong Eo$^1$\thanks{$^*$ Equal Contribution}  , Hyeonseok Moon$^{1*}$, Jinsung Kim$^{1*}$, Yuna Hur$^{1*}$, Jeongwook Kim$^{1*}$ \\ \textbf{Songeun Lee$^{2}$, Changwoo Chun$^{2}$, Sungsoo Park$^{2}$, Heuiseok Lim$^1$\thanks{$^\dagger$ Corresponding Author}}
 \\
$^1$Dept. of Computer Science and Engineering, Korea University  \\$^2$Hyundai Motor Group \\
\texttt{\{djtnrud,limhseok\}@korea.ac.kr} \quad \texttt{songeun.lee@hyundai.com}}

\begin{document}
\maketitle

\begin{abstract}
Recent advances in QA pair generation (QAG) have raised interest in applying this technique to the educational field. However, the diversity of QA types remains a challenge despite its contributions to comprehensive learning and assessment of children. In this paper, we propose a QAG framework that enhances QA type diversity by producing different interrogative sentences and implicit/explicit answers. Our framework comprises a QFS-based answer generator, an iterative QA generator, and a relevancy-aware ranker. The two generators aim to expand the number of candidates while covering various types. The ranker trained on the in-context negative samples clarifies the top-N outputs based on the ranking score. Extensive evaluations and detailed analyses demonstrate that our approach outperforms previous state-of-the-art results by significant margins, achieving improved diversity and quality. Our task-oriented processes are consistent with real-world demand, which highlights our system's high applicability. Our code is available at \url{https://github.com/sugyeonge/Towards-diverse-QAG.git}.
\end{abstract}

\section{Introduction}
Pedagogical studies over the years have demonstrated that asking questions about a given storybook nurtures insight and expands knowledge~\cite{janusheva2009questions,etemadzadeh2013role,shanmugavelu2020questioning}. 
Hence, posing questions becomes a fundamental part of education to engage children and promote literacy~\cite{cotton1988classroom, ellis1993teacher, dillon2006effect}.
Along with the remarkable strides in natural language processing, recent studies have actively explored question-answer pair generation (QAG) systems that target education~\cite{xu-etal-2022-fantastic,yao-etal-2022-ais,zhao-etal-2022-educational}. As QAG is a labor-intensive manual process, it benefits from automated production methods. Furthermore, sustainable system update and utilization emphasize their high applicability~\cite{le2014automatic,jerome2021automatic}.

A challenge in educational QAG is the diversity of generated QA pairs as well as their quality~\cite{lee-etal-2020-generating,zhang2021review}. Exploiting various QA types facilitates comprehensive learning, as each question inquires information specific to its type and stimulates different brain activities in the answering process~\cite{guszak1967teacher,dillon2006effect}.
Controlling difficulty by adopting different types of questions or answers also enables a balanced assessment of the reading comprehension skills of children~\cite{xu-etal-2022-fantastic}. Consequently, actively using questions with various interrogative words and answers reflecting both implicitness and explicitness is important.
Yet, existing educational QAG studies have rarely considered diversity. Generated questions of existing models are extremely biased to the `What' and `Who' type questions. Answer extraction focuses on detecting spans within passages, resulting in an inability to create implicit answers that do not directly appear in the passage.

To address the limitation, we propose an effective QAG framework that enhances diversity and quality. Our framework consists of a QFS-based answer generator, an iterative QA generator, and a relevancy-aware ranker.
Specifically, \textbf{QFS-based answer generator} adopts query-focused summarization (QFS)-based~\cite{vig-etal-2022-exploring} answer generation model (AGM), with the aim of obtaining diverse and proper answer candidates.
\textbf{Iterative QA generator} is designed to increase question type variety by exploiting the interrogative word-indicated question generation model (QGM). We jointly execute this QGM with the question-answering model (QAM) to adjust the final answers. \textbf{Relevancy-aware ranker} inspects quality to determine the final top-N outputs among the generated candidates. To grasp better pairs with high relevancy, the ranker is trained using in-context negative samples.

\begin{figure*}[]
\centering
\includegraphics[width=\linewidth]{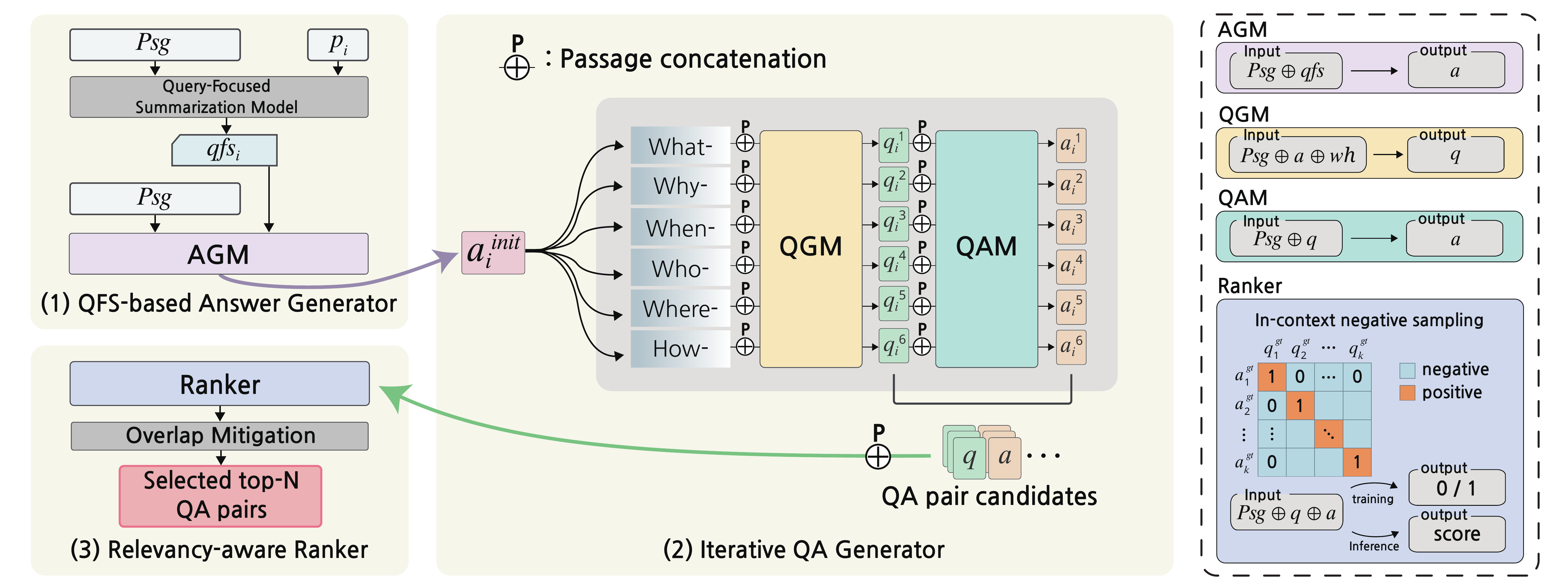}
 \caption{Overall architecture of our QAG framework. The rightmost side on the figure describes the training process of each model.} 
 \label{fig:main}
\end{figure*}

The experimental results indicate that our framework outperforms the existing state-of-the-art method by a large margin, with a gain of up to 0.435$\rightarrow$0.503 on MAP@N with Rouge-L f1 and 0.9077$\rightarrow$0.9178 on MAP@N with BERTScore~\cite{yao-etal-2022-ais}. Additional statistical and human evaluations with detailed analyses consistently show higher QA type diversity and quality compared to previous studies, which demonstrates the superiority of the proposed approach. The three modules of our framework are process-oriented, providing outputs from each, which is in line with the real-world demand investigated by \citet{wang2022towards}. This highlights the high applicability to the education field in terms of Human-AI collaboration.
We summarize our contributions as follows: 
\begin{enumerate}[label=(\roman*)]
    \item We propose a novel QAG framework that enhances the diversity of question and answer types while increasing quality. 
    \item Extensive experiments show that our framework remarkably outperforms previous state-of-the-art results with high diversity and relevance. 
    \item The task-oriented process is consistent with real-world demand, emphasizing the applicability of our framework in the education field.
\end{enumerate}

\section{Related Works}
The question-answer pair generation (QAG) task aims to automatically generate QA pairs based on the input text.
In the early days, rule-based QAG systems are dominant~\cite{lindberg2013generating,labutov2015deep}. With the advent of a deep learning-based paradigm, it was demonstrated for the first time by \citet{du2017learning} that a fully end-to-end QAG system generates exceptionally good questions. Accordingly, diverse studies have been conducted and developed~\cite{shakeri-etal-2020-end,li2022consecutive,zhou-etal-2019-multi}.
\citet{kang2019let} adopt an interrogative words-based approach to clarify the semantics of words from a passage, resulting in the generation of questions containing key information of the context.
\citet{scialom2019self} attempt to generate questions in an answer-agnostic manner by adapting the self-attention mechanism of Transformers with a copying mechanism, placeholders, and contextual word embeddings.
\citet{dong2022closed} propose a QAG model for closed-book setting without access to external knowledge by modeling the semantic relationships between questions and answers at a contextual level and measuring the answerability of the generated questions.

In recent times, several attempts have been made to automatically generate valid QA pairs for educational purposes.
FairytaleQA proposed by \citet{xu-etal-2022-fantastic} is a representative dataset in educational QAG. Education experts manually generated QA pairs suitable for learning and assessing children's reading comprehension skills. With the dataset, \citet{yao-etal-2022-ais} present an educational QAG system through a combination of three-step modules. \citet{zhao-etal-2022-educational} deal with the high-cognitive demand question generation based on three out of seven narrative elements in the FairytaleQA. \citet{dugan-etal-2022-feasibility} summarize QA pairs to given book chapters and provide them to the fine-tuned T5~\cite{raffel2020exploring} models. 

However, these studies rarely consider diversity when performing QAG. Leveraging diverse QA types are important aspect in QAG since using various interrogative words promotes different parts of the brain, which facilitates children's comprehensive learning~\cite{guszak1967teacher,dillon2006effect}. 
Varying answer types is also a factor that contributes to a balanced assessment, as the difficulty can be controlled by adjusting whether the answer is revealed in the passage~\cite{xu-etal-2022-fantastic}.
The evidence emphasizes the importance of considering a variety of QA pairs from a broad perspective for the effectiveness of reading comprehension~\cite{kim2017simple}.

\section{Method}
Our QAG framework comprises three task-oriented processes: a QFS-based answer generator, an iterative QA generator, and a relevancy-aware ranker. The main goal of the two generators is to expand QA pair candidates containing diverse question and answer types. The ranker aims to determine the final output by scoring QA pair candidates. The overall QAG architecture of our framework is depicted in Figure \ref{fig:main}.

\subsection{QFS-based Answer Generator}
In the initial answer generation process, we employ query-focused summarization (QFS) to capture salient information related to a given sentence. After the QFS model generates a query-focused summary of a given passage by referring to the relevant key, the summary is fed into the generative answer generation model (AGM) to output implicit or explicit answers.

Let $\textit{Psg}$ denote a passage consisting of $n$ sentences, $p_{1}, ..., p_{n}$, and the corresponding ground-truth (GT) QA pair be $(Q^{\text {gt}}, A^{\text{gt}}) = {\{(q^{\text{gt}}_{j}, a^{\text{gt}}_{j})\}}_{j=1}^{m}$. 
First, we generate a query $q^{\text{gt}}_{j}$ focused summary $\textit{qfs}^{\text{gt}}_{j} = \textit{QFS}(\textit{Psg}, q^{\text{gt}}_{j})$ of $\textit{Psg}$ using the pre-trained QFS model, $\textit{QFS}$.
We then train AGM, termed as $\theta_{\text{AGM}}$, with the concatenated input of $\textit{Psg}$ and $\textit{qfs}^{\text{gt}}_{j}$ in a sequence-to-sequence manner. The loss function for each $\textit{Psg}$ is estimated as shown in Equation~(\ref{eq:3.1.initAG}).

\begin{equation} \label{eq:3.1.initAG} \footnotesize
    \begin{split}
 L_{\text{AGM}} = - \displaystyle\sum_{(q^{\text{gt}}_{j}, a^{\text{gt}}_{j}) \in (Q^{\text{gt}}, A^{\text{gt}})} \mathbf{E}_{\theta_{\text{AGM}}}(a^{\text{gt}}_{j} \mid \textit{Psg}, \textit{qfs}^{\text{gt}}_{j})
    \end{split}
\end{equation}

In the inference phase, for each sentence $p_{i}$ in $\textit{Psg}$, we generate $\textit{qfs}_{i} = \textit{QFS}(\textit{Psg},p_{i})$. Then AGM produces a single initial answer $a_{i}^{init}$ for corresponding $\textit{qfs}_{i}$. 
The resulting answer set $A^{init}$ has $n$ answers since answers are generated for every sentence in the passage. $A^{init}$ is expressed as follows:

\begin{equation} 
    A^{init} = \{ \theta_{\text{AGM}}(\textit{Psg}, \textit{qfs}_{i}) \mid p_{i} \in \textit{Psg} \}
\end{equation}

\subsection{Iterative QA Generator} 
After the initial answer set $A^{init}$ is generated, the next step is to expand the QA pair candidates to reflect the question type diversity. To achieve this, we propose an interrogative word-indicated question generation model (QGM), denoted by $\theta_{\text{QGM}}$, and a generative question-answering model (QAM), denoted by $\theta_{\text{QAM}}$. The QGM and QAM are sequentially executed based on the initial answer to generate a set of QA pair candidates. The following paragraphs describe the training and inference processes of each model.

\paragraph{Interrogative word-indicated QGM}

We train QGM with GT QA pair set to generate questions by referring to the answers and their passages. Including interrogative words in the training phase allows controllable question generation to follow the desired interrogative type during inference.

We denote the interrogative word of each $q^{\text{gt}}_{j}$ in a GT QA pair set as $wh^{\text{gt}}_{j}$. In our setting, $wh$ is an element of the interrogative word set $\textit{WH}$=$\{$Who, When, What, Where, Why, How$\}$. $\theta_{QGM}$ is trained to generate question $q^{\text{gt}}_{j}$ by feeding the concatenated input of $Psg$, $a^{\text{gt}}_{j}$, and $wh^{\text{gt}}_{j}$. Training is performed in a sequence-to-sequence manner and is optimized using the following loss function:

\begin{equation} \footnotesize
    \begin{split}
 L_{\text{QGM}} = - \displaystyle\sum_{(q^{\text{gt}}_{j}, a^{\text{gt}}_{j}) \in (Q^{\text{gt}}, A^{\text{gt}})} \mathbf{E}_{\theta_{\text{QGM}}}(q^{\text{gt}}_{j} \mid \textit{Psg}, a^{\text{gt}}_{j}, wh^{\text{gt}}_{j})
    \end{split}
\end{equation}

In the inference phase, we prioritize diversity and generate questions by considering each interrogative word in $\textit{WH}$ as an indicator.
For each $a_{i}^{init} \in A^{init}$ generated in the first step and its corresponding passage \textit{Psg}, $\theta_{\text{QGM}}$ configures QA pair set ${QA}^{1}$ which can be expressed as follows:

\begin{equation} \footnotesize
    \begin{split}
        {QA}^{1} = \{ \textbf{(} \theta_{\text{QGM}}(\textit{Psg}, a^{init}_{i}, wh), a^{init}_{i} \textbf{)} \mid wh \in \textit{WH}, \\ 
        \quad \quad \quad a_{i}^{init} \in A^{init}\}
    \end{split}
\end{equation}

In this way, QA pair candidates with high relevance to the passage can be generated. Note that this process encourages the expansion of question types, not all questions generated are related to the initial answers.

\paragraph{Answer Adjustment}
To consider relevancy between QA pairs, we reconstruct answers through $\theta_{\text{QAM}}$ trained with a set of GT QA pairs. 
This process helps avoid linking inappropriate questions to a given initial answer, such as asking a `How' question for an answer aimed at a specific person. Training of $\theta_{\text{QAM}}$ is proceeded by optimizing the following loss function.

\begin{equation} \footnotesize  \label{eq:3.2.QAM}
    \begin{split}
 L_{\text{QAM}} = - \displaystyle\sum_{(q^{\text{gt}}_{j}, a^{\text{gt}}_{j}) \in (Q^{\text{gt}}, A^{\text{gt}})} \mathbf{E}_{\theta_{\text{QAM}}}(a^{\text{gt}}_{j} \mid \textit{Psg}, q^{\text{gt}}_{j})
    \end{split}
\end{equation}

In the subsequent inference phase, we adjust the answers to all questions in ${QA}^{1}$ through $\theta_{\text{QAM}}$. The reconstructed QA pair set, denoted by ${QA}^{2}$, is expressed as Equation~(\ref{eq:3.2.QA2}).

\begin{equation} \footnotesize \label{eq:3.2.QA2}
     {QA}^{2} =  \{ \textbf{(} q^{j}_{i}, \theta_{\text{QAM}}(\textit{Psg},q^{j}_{i}) \textbf{)} \mid (q^{j}_{i}, a^{init}_{i}) \in {QA}^{1}\}
\end{equation} 

${QA}^{2}$ is a final QA pair candidate set in which the relevance between the pairs is supervised through the QAM while maintaining the diversity of question types.

\subsection{Relevancy-aware Ranker} 
With the relevancy-aware ranker model, we select top-N ranked QA pairs that exhibit high relevance between passages and QA pairs. 

The ranking model denoted by $\theta_{Rank}$ produces the relevance score for each QA pair. To train the ranking model $\theta_{Rank}$, we compose a contrastive training dataset by collecting in-context negative samples in GT QA pair set. In the training data, the GT QA pairs are considered a positive samples, and the other QA pairs within the same passage are considered negative samples. For a given passage \textit{${Psg}$} and the corresponding GT QA pair set $(Q^{\text{gt}}, A^{\text{gt}})$, we construct positive sample set $\textit{POS} = \{ (q^{\text{gt}}_{i}, a^{\text{gt}}_{j}) \mid q^{\text{gt}}_{i} \in Q^{\text{gt}}, a^{\text{gt}}_{j} \in A^{\text{gt}}, i = j\}$ and negative sample set $\textit{NEG} = \{ (q^{\text{gt}}_{i}, a^{\text{gt}}_{j}) \mid q^{\text{gt}}_{i} \in Q^{\text{gt}}, a^{\text{gt}}_{j} \in A^{\text{gt}}, i \neq j\}$\footnote{We consider QA pairs in a different passages as easy negative cases and do not include them as negative samples in the ranker training.}.

Then the QA pairs and their corresponding passages are concatenated to construct the input sequences for training $\theta_{Rank}$. By feeding this input sequence, $\theta_{Rank}$ is trained to classify binary labels representing negative and positive.

In the inference phase, $\theta_{Rank}$ returns the scores of the input QA pair to be classified as positive and negative, respectively. We further rank each QA pair by referring both scores.

Through this process, the ranker is trained to prioritize the selection of data that exhibits a high correlation between QA pairs and high relevance to the corresponding passages.

\paragraph{Overlap Mitigation}
While the ranker model enhances the relevance of QA pairs, the issue of duplication exists where the top-ranked pairs constitute similar forms. To alleviate this issue, we compute a re-scaled ranking score to diminish the lexical overlap of answers in the QA pair candidates.

We sequentially select QA pairs in the order of high scores computed using the ranking model. 
To consider lexical overlap in each selection process, we measure the Rouge-L score between the selecting pair and the previously selected QA pairs. 
The score $s$ of each pair measured by the ranking model is re-scaled as $s-{Rouge}* abs(s)$.
Through this process, we down-scale the scores of the QA pairs that exhibit high lexical overlap with previously selected QA pairs. This allows the selection of various types of QA while reflecting the scores calculated by the ranking model. The detailed procedure of the overlap mitigation algorithm is presented in Algorithm~\ref{alg:overlap} in Appendix~\ref{app:overlap_algorithm}.

\section{Experiments}
\subsection{Experimental Setup}
\begin{table*}[]
\resizebox{\textwidth}{!}{
\centering
\begin{tabular}{c|cccc|cccc}
\toprule[1.5pt]
{} & \multicolumn{4}{c|}{\textbf{MAP@N (Rouge-L F1)}} & \multicolumn{4}{c}{\textbf{MAP@N (BERTScore F1)}} \\ \midrule
\textbf{Method} & \textbf{Top 10}& \textbf{Top 5} & \textbf{Top 3} & \textbf{Top 1} & \textbf{Top 10}  & \textbf{Top 5}& \textbf{Top 3}& \textbf{Top 1}\\ \midrule[1.5pt]
\makecell[l]{\textsl{FQAG}\cite{yao-etal-2022-ais}} & 0.440 / 0.435& 0.375 / 0.374& 0.333 / 0.324& 0.238 / 0.228& 0.9077 / 0.9077 & 0.8990 / 0.8997& 0.8929 / 0.8922& \textbf{0.8768} / 0.8776 \\\hline
\makecell[l]{\textsl{SQG}\cite{dugan-etal-2022-feasibility}} & 0.460 / 0.455& 0.392 / 0.388& 0.344 / 0.337& 0.234 / 0.242& 0.9056 / 0.9062& 0.8953 / 0.8955& 0.8876 / 0.8878& 0.8707 / 0.8723\\\hline 
\makecell[l]{Ours} & \textbf{0.500 / 0.503} & \textbf{0.426 / 0.429} & \textbf{0.369 / 0.372}& \textbf{0.247/ 0.254} & \textbf{0.9156 / 0.9178} & \textbf{0.9046 / 0.9068} & \textbf{0.8956 / 0.8977}& 0.8752 / \textbf{0.8783}\\
\bottomrule[1.5pt]
\end{tabular}}\caption{The main experimental results for our QAgen framework. We report Map@N score with Rouge-L F1 and BERTScore F1 for each model. The result for the validation split is on the left side, and the right side is for the test split. 
}\label{tb:main_result}
\end{table*}
\begin{table*}[]
\centering
\resizebox{\textwidth}{!}{
\begin{tabular}{c|ccc||ccccc}
\toprule[1.5pt]
  {} & \multicolumn{3}{c||}{\textit{global}} & \multicolumn{5}{c}{\textit{local}} \\ \midrule
  \textbf{Method} & \textbf{Diversity-Q $\downarrow$}  & \textbf{Diversity-A $\downarrow$} & \textbf{Quality-E$\downarrow$} & \textbf{Relevancy $\uparrow$} & \textbf{Acceptability $\uparrow$}  & \textbf{Usability $\uparrow$} & \textbf{Readability $\uparrow$}   & \textbf{Difficulty $\uparrow$} \\ \midrule[1.5pt]
\makecell[l]{\textsl{FQAG}\cite{yao-etal-2022-ais}} & 3.03 & 3.06 & 2.66 & 2.65 & 2.14 & 1.74 & \textbf{2.64} & 1.11 \\ \hline
\makecell[l]{\textsl{SQG}\cite{dugan-etal-2022-feasibility}} & 2.96 & 3.03 & 3.30 & 2.44 & 1.87 & 1.34 & 2.55 & 1.36 \\ \hline
\makecell[l]{Ours}& \textbf{2.35} & \textbf{2.18} & \textbf{2.35} & \textbf{2.69} & \textbf{2.22} & \textbf{1.9} & 2.35 & \textbf{1.98} \\ \hline \hline
\makecell[l]{GT} & 1.65 & 1.71  & 1.68 & 2.97 & 2.65 & 2.50 & 2.80 & 1.95 \\ \bottomrule[1.5pt]
\end{tabular}}
\caption{Human evaluation results for the QA pairs generated by the QAG systems on eight criteria. \textit{global} represents the human ranking results for the three QAG systems and GT. \textit{local} indicates the human scoring results for each QAG system and GT, on a 0-3 scale. Note that the scores between the two settings are completely different.
}\label{tb:human}
\end{table*}

\paragraph{Dataset}
In our experiments, we leverage the FairytaleQA dataset~\cite{xu-etal-2022-fantastic}. FairytaleQA is specifically designed for children's storybook learning and assessment, which corresponds to our purpose of education. In the data construction process, educational experts manually created QA pairs to ensure reliability and validity. The training, validation, and test sets contain 8,548 QA pairs from 232 books, 1,025 pairs from 23 books, and 1,007 pairs from 23 books, respectively. Instead of using narrative elements (\textit{i.e.} character, setting, action, etc.) presented in the dataset, we diversify the questions based on interrogative words to induce expanded types of questions beyond these elements. We use the existing answer types, as they are mutually exclusive.

\paragraph{Models}
All models comprising our framework are trained with the FairytaleQA dataset. In the case of the QFS model, we produce a summary using model checkpoints provided by \citet{vig2021exploring}. In training AGM, QGM, and QAM, we exploit the pre-trained BART-large~\cite{lewis2020bart} model and framework provided by Fairseq\footnote{\textcolor{blue}{\url{https://github.com/facebookresearch/fairseq.git}}}. For the hyper-parameters, 2048 max tokens, early stopping 10, and polynomial decay scheduler are adopted. For the learning rate and dropout, we set 3e-05 and 0.1 in AGM and QGM, 2e-05 and 0.2 in QAM, respectively. All models are trained on 2 RTX8000 GPUs. We used the RoBERTa-base~\cite{liu2019roberta} model and Huggingface\footnote{\textcolor{blue}{\url{https://github.com/huggingface/transformers}}} framework for our ranking model. We train it for five epochs with a fixed learning rate of 5e-07 and a single GPU is used for training ranker.

\subsection{Evaluation Metrics}
For the evaluation metric, we adopt the MAP@N score as a primary metric utilized by \citet{yao-etal-2022-ais}. MAP@N with Rouge-L refers to the averaged value of the maximum score set added by computing Rouge-L between each GT pair and the top-N generated QA pairs. Each question and answer in the QA pair is concatenated in the process. However, when MAP@N is measured by the Rouge-L precision score as in \citet{yao-etal-2022-ais}, short results are advantageous. This is because it measures the longest overlap over the number of candidates. Instead of using precision, we select the F1 score for accurate measurement.
Since the metrics based on the n-gram overlap do not guarantee quality~\cite{zhang2019addressing}, we additionally adopt BERTScore for MAP@N to evaluate semantic equivalence based on similarity scores~\cite{zhang2019bertscore}.

\subsection{Baselines}
We adopt two educational QAG systems as a baseline.

\paragraph{\textsl{FQAG}} \textsl{FQAG}~\cite{yao-etal-2022-ais} is a state-of-the-art study of FairytaleQA. They perform QAG through a three-step pipeline comprising answer generation, question generation, and ranking modules. For re-implementation, we load the provided checkpoints to generate QA pairs for the validation and test sets of FairytaleQA.
\paragraph{\textsl{SQG}} \textsl{SQG}~\cite{dugan-etal-2022-feasibility} is a recently published paper in educational QAG, which utilizes summaries of the given passages. QA pairs are generated leveraging three models: answer generation, question generation, and question answering models. In this case, to match the number of top-N, we select QA pairs based on the generated order or increase outputs by adjusting the beam size.

\section{Results and Analysis}

\subsection{Automated Evaluation}
\paragraph{Result on MAP@N with Rouge-L}

\begin{figure}[]
\centering
\includegraphics[width=\linewidth]{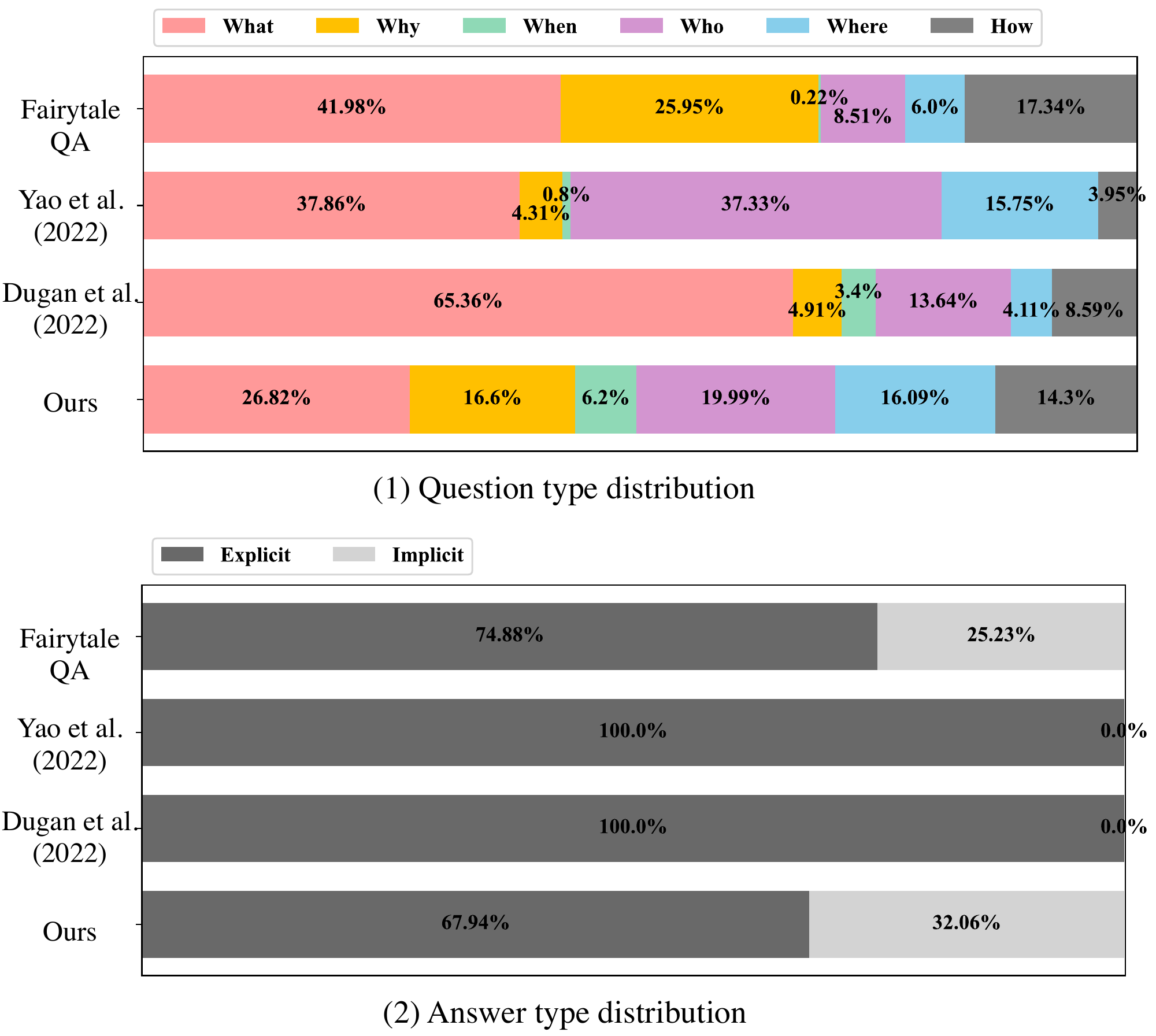}
 \caption{Distribution of the question and answer types on the test set. It represents the portion of what, why, when, who, where, and how from left to right.}

 \label{fig1:data_stat}
\end{figure}

Table \ref{tb:main_result} shows the main result of MAP@N with Rouge-L F1 scores according to the QAG systems. As a result, our system significantly outperforms the baseline model in all splits and top-N outcomes. 
Especially in the test set, we outperform \textsl{FQAG} by +0.068 in the top 10, +0.055 in the MAP@5, and +0.048 in the MAP@3, which is a significant gain. \textsl{SQG} achieves better results than \textsl{FQAG} but still does not outperform ours. Compared to the \textsl{SQG}, our system shows improvement in all top-N results mainly from 0.455 to 0.503 (+0.048).
The result implies that generating various QA pair candidates and properly establishing plausible pairs serve as one contributing factor to performance improvement. 

\paragraph{Result on MAP@N with BERTScore}
We measure MAP@N by employing BERTScore to evaluate the semantic equivalence between GT and generated QA pairs. Namely, we use the F1 value of BERTScore instead of Rouge-L F1 score when measuring MAP@N.
As a result, our system achieves higher performance in all settings except for the MAP@1 validation result. In the best case from the test result (MAP@10), \textsl{FQAG} and \textsl{SQG} showed 0.9077 and 0.9062 respectively and we recorded 0.9178, outperforming by +0.0101 and +0.0116. The tendency for our performance to be the highest is consistent with the MAP@N with Rouge-L F1 result. However, we observe that \textsl{FQAG} reports higher performance in BERTScore than \textsl{SQG}. Although the performance gap is marginal, this outcome suggests that the generated QA pairs of \textsl{FQAG} are semantically better than \textsl{SQG}.

\subsection{Statistical Evaluation}
To evaluate the question and answer type diversity of generated QA pairs, we perform a statistical evaluation. The distribution according to interrogative type and answer type is presented in Figure \ref{fig1:data_stat}. As a result, the reported question types of ours are more balanced than others. Unlike other models that usually create `what' and `who', our QAG system is well balanced with questions of `why' and `how' that require reasoning.
This suggests the potential of children to think from various perspectives by being asked different types of questions.

For answer types, our system contains 32.06\% of implicit answers, indicating that implicit answers are also well generated, which allows our model to help balance assessments of children. Conversely, other models use the answer span extraction method, resulting in a 0\% of implicit answers.

\subsection{Human Evaluation}
We further conduct a human evaluation for a detailed inspection. For each paragraph, three human evaluators with degree holders or experts in education rate each of the three QA pairs generated by the GT and three QAG systems. Human evaluation is performed on a total of 20 passages, and we select three QA pairs sequentially for unscored GT and \textsl{SQG}.
Due to the brevity of space, we further describe human evaluation details in the Appendix \ref{app:human_eval}.
The following criteria are used for human evaluation. For the \textit{global} setting, we instruct the evaluator to rank the entire system, and in the \textit{local} cases to select how many of the three QA pairs generated by each system correspond to property items.

(\textit{global} setting) \textbf{Diversity-Q}: This ranks the generation results of GT and three QAG systems in terms of question diversity.
\textbf{Diversity-A}: This ranks the generation results of GT and the three QAG systems in terms of answer diversity.
\textbf{Quality-E}: This ranks the entire system quality from an overall perspective.

(\textit{local} setting) \textbf{Relevancy}: This evaluates the relevance between a passage and a QA pair. If either question or answer is not relevant, it is irrelevant.
\textbf{Acceptability}: This evaluates whether a question and its corresponding answer are correctly generated. Relevance with the passage is not considered, and if either of them is awkward, it is considered incorrect.
\textbf{Usability}: This evaluates whether the generated QA pairs can be used for education purposes.
\textbf{Readability}: This evaluates whether the generated QA pairs are grammatically right.
\textbf{Difficulty}: This evaluates whether the generated QA pairs are excessively easy.

Table \ref{tb:human} presents the results of the human evaluation. 
Our approach achieves remarkable performance in terms of the question and answer diversity with an average ranking of 2.35 and 2.18 respectively. In the global setup, we observe that the Quality-E is 2.66 in \textsl{FQAG} and 3.30 in \textsl{SQG}, while our system outperforms them with a score of 2.35. These results demonstrate that our QAG is both quantitatively and qualitatively superior in direct comparison with other systems through ranking while enhancing diversity.
The results of the local setting indicate that we outperform both \textsl{FQAG} and \textsl{SQG} except for the readability. 
As an evaluation result of the QA pairs we generated, the relevance of the passages to the generated QA (2.69), the acceptance of the questions to the answers (2.22), and the usability for educational purposes (1.9) show the highest result compared to other systems. We even observe a slight performance gain over GT in case of difficulty. However, in readability, our result showed 2.35, which is lower than the 2.64 and 2.55 of the existing model. We speculate that the average length of our generated QA pairs is longer, resulting in a small trade-off with difficulty. From the results, we conclude that our generated QA pairs are truly effective in not only ensuring quality but also diversity.

\subsection{Ablation Study}
We perform ablation studies to further analyze the contribution of each process in our framework to the overall performance. The results are shown in Table \ref{tb:ablation}. 

\paragraph{Impact of Query-focused Summarization}
Instead of the AGM model, we generate answers using noun phrase and noun entity extraction method performed in \textsl{FQAG}. When the AGM model is changed, the performance decreased by -0.031 in MAP@10. This indicates that introducing a summary containing intensive information benefits from generating more plausible answers. 
We also analyze that introducing the AGM model in a generative manner yields higher performance because it is capable to generate implicit answers.

\begin{table}[]
\centering
\resizebox{0.48\textwidth}{!}{
\begin{tabular}{c|cccc}
\toprule[1.5pt]
{} & \multicolumn{4}{c}{\textbf{MAP@N (Rouge-L F1)}}\\ \midrule
\textbf{Method} & \textbf{Top 10} & \textbf{Top 5} & \textbf{Top 3} & \textbf{Top 1} \\ \midrule[1.5pt]
\makecell[l]{Ours} & 0.503 & 0.429& 0.372& 0.254\\ \hline \hline
\makecell[l]{w/o QFS} & 0.472 & 0.401 & 0.348 & 0.248\\ \hline
\makecell[l]{w/o Iteration}   & 0.463 & 0.427& 0.378& 0.253\\ \hline
\makecell[l]{w/o Contrastive learning} & 0.438 & 0.375& 0.326& 0.261 \\ \bottomrule[1.5pt]
\end{tabular}}
\caption{Ablation results on the test set to claim the necessity of each module. Every module functions in \textit{Ours}.}\label{tb:ablation}
\end{table}
\paragraph{Impact of Iterative QA Pair Generation}
For investigating the most effective iteration, we reduce the number of iterations in the iterative QA generator. Namely, we eliminate the QAM step and execute only QGM. The experimental results show a marginal change in most of the cases, except for the top 10 results. We attribute this performance drop in the top 10 to the process of additionally adjusting interrogative word-indicated questions to correct answers through QAM. An experiment on increasing iterations is presented in Section \ref{sec:iter}.

\paragraph{Impact of Contrastive Learning}
To observe the role of our relevancy-aware ranker, we eliminate our ranker and utilize the DistilBERT~\cite{sanh2019distilbert} ranking model of \textsl{FQAG}.
As a result of the experiment, the overall performance is degraded largely, such as 0.503$\rightarrow$0.438 in the MAP@10 and 0.429$\rightarrow$0.375 in the MAP@5. We interpret that constituting the training examples for contrastive learning through in-context negative samples boosts the overall performance gain.

However, our performance of changing the ranking model to that of \textsl{FQAG} can also be compared with the \textsl{FQAG} test result of Table~\ref{tb:main_result} (MAP@10: 0.435, MAP@5: 0.374, MAP@3: 0.324, MAP@1: 0.228). This case is a comparison in which the ranking model is unified and only varies the QA pair generation part.
Results show an insignificant difference, with \textsl{FQAG} test results in Table~\ref{tb:main_result} performing lower than ablation results of contrastive learning. We analyze that our method generates more QA pair candidates with the goal of increasing diversity, but the DistilBERT ranking model does not rank them well.

\section{Case Study}

\paragraph{Performance of Multiple AGM}
We investigate various methods to add a clue that can be a key element when constructing AGM inputs. The clue is then fed into the BART large-based AGM input along with the given passages, and the answer is predicted. \textit{A} is the baseline where this learns to directly generate answers for given passages. 
In \textit{DS}, one to three sentences in the passage closest to the question are retrieved. In \textit{Ext-Ret}, a phrase or sentence closest to the question is retrieved from the external resource NarrativeQA~\cite{kovcisky2018narrativeqa}.

\begin{table}[h]
\centering
\resizebox{0.5\textwidth}{!}{
\begin{tabular}{l|ccc}
\toprule[1.5pt]
 \multicolumn{1}{c|}{\textbf{Method}} & \multicolumn{1}{c}{\textbf{Rouge-L}} & \multicolumn{1}{c}{\textbf{BLEU}} & \multicolumn{1}{c}{\textbf{BERTScore}} \\ \midrule[1.5pt]
A  & 0.216   & 8.31 & 0.875\\ \hline
DS1 & 0.232   & 10.21& 0.879\\ \hline
DS2& 0.244   & 9.65 & 0.874\\ \hline
DS3& 0.256   & 10.55& 0.878\\ \hline
Ext-Ret (Sent)   & 0.283   & 14.86& 0.89 \\ \hline
Ext-Ret (Phrase) & 0.304   & 16.74& 0.896\\ \hline
\textbf{QFS}     & \textbf{0.362}& \textbf{23.21}           & \textbf{0.903} \\ \bottomrule[1.5pt]
\end{tabular}
}
\caption{FairytaleQA test set evaluation results according to the answer generation model}\label{tb:ag_model}
\end{table}

Table~\ref{tb:ag_model} is the experimental result, and the performance of QFS outperforms all other methodologies. For this result, we judge that the summary, in which the information is compressed and regenerated, contributes more to the final answer generation.

\paragraph{Performance on Adding Iteration} \label{sec:iter}
We observe the performance fluctuation when increasing iteration on the iterative QA generator. We create QA pairs by recursively executing QGM and QAM on the QA pairs generated by our main framework. Experimental results in Table~\ref{tb:iter} show that the performance degrades as the iteration increases. We judged that no additional performance improvement would be obtained even if iterations were repeated more than this.
\begin{table}[h]
\centering
\resizebox{0.5\textwidth}{!}{
\begin{tabular}{c|c|c|c|c}
\toprule[1.5pt]
{} & \multicolumn{4}{c}{\textbf{Map@N (Rouge-L F1)}}\\\midrule
\textbf{Method} & \multicolumn{1}{c|}{\textbf{Top 10}} & \multicolumn{1}{c|}{\textbf{Top 5}} & \multicolumn{1}{c|}{\textbf{Top 3}} & \multicolumn{1}{c}{\textbf{Top 1}} \\ \midrule[1.5pt] 
\multicolumn{1}{l|}{Ours} & 0.503 & \textbf{0.429}& \textbf{0.372}& \textbf{0.254}\\ \hline \hline
\multicolumn{1}{l|}{Ours +1 iteration} & \textbf{0.506}& 0.423& 0.361& 0.246\\ \hline
\multicolumn{1}{l|}{Ours +2 iteration} & 0.502 & 0.419& 0.362& 0.243\\ \bottomrule[1.5pt]
\end{tabular}}\caption{Result on iteration. +1 iteration refers to additionally attaching a QGM model. +2 refers to successively applying the QAM model.}\label{tb:iter}
\end{table}

\paragraph{Performance on Overlap Mitigation Methods}
\label{app:overlap_algorithm}
In this section, we investigate the effect of overlap mitigation techniques. \textit{EM} is a baseline, which remains the highest-scored QA pair for each unique \textit{Criterion}.

The experiment is designed to modify two factors: In \textit{Criterion}, we divide the criterion of overlap measurement into two parts of question or answer.
\textit{Overlap Metric} divides the overlap measurement metric into BLEU and Rouge.

\begin{table}[h]
\centering
\resizebox{0.5\textwidth}{!}{
\begin{tabular}{c|l|cccc}
\toprule[1.5pt]
\multirow{2}{*}{\textbf{Criterion}} & \multirow{2}{*}{\makecell{\textbf{Overlap} \\ \textbf{Metric}}} & \multicolumn{4}{c}{\textbf{MAP@N (Rouge-L F1)}}\\ \cmidrule{3-6}
{} & {} & \multicolumn{1}{c}{\textbf{Top 10}} & \multicolumn{1}{c}{\textbf{Top 5}} & \multicolumn{1}{c}{\textbf{Top 3}} & \multicolumn{1}{c}{\textbf{Top 1}} \\ \midrule[1.5pt]
\multirow{3}{*}{Answer} & EM & 0.491& 0.414& 0.357& 0.254\\ \cline{2-6}
{} & BLEU  & 0.497& \textbf{0.431}& 0.369& 0.254\\ \cline{2-6}
{} & Rouge-L & \textbf{0.503}& 0.429& \textbf{0.372}& 0.254\\ \midrule
\multirow{3}{*}{Question} & EM & 0.483& 0.404& 0.354& 0.254\\ \cline{2-6}
{} & BLEU  & 0.491& 0.421& 0.365& 0.254\\ \cline{2-6}
{} & Rouge-L & 0.\textbf{493}& \textbf{0.431}& 0.\textbf{366}& 0.254\\ \bottomrule[1.5pt]
\end{tabular}
}\caption{Experimental results for various overlap mitigation methods. The Top 1 score for the overall models is the same since the QA pair with the highest score is always selected first.}\label{tb:overlap}
\end{table}

The experimental results are presented in Table~\ref{tb:overlap}. The results represent that re-scaling the scores of the ranking model by using overlap mitigation methods yields higher performance than the method of simply removing overlap based on exact matching. Also, overall performance shows better when the overlap metric is set to Rouge than BLEU.
This demonstrates that the output of the ranking model can be utilized more effectively by applying the proposed overlap mitigation method. Notably, the overlap mitigation method based on the answer record higher performance when the question is used as the criterion. 

\section{Conclusion}
In this paper, we proposed a QAG framework for educational purposes featuring diverse and valid question and answer types.
Our framework is structured with three task-oriented processes, with a particular emphasis on expanding diverse and valid types of QA pair candidates in the generator, and selecting high-quality QA pairs in the ranker. We conducted extensive evaluations of generated QA pairs, including quantitative, qualitative, and statistical evaluations with detailed analyses, and observed that our system achieved remarkable performance. Our framework has the potential to promote various cognitive activities in children learning by providing diverse and effective QA pairs for educational purposes. As our modularized task-oriented frameworks are tailored to real-world demand, we further expect the collaborative use of humans and AI.

\section*{Limitations}
We used only the pre-trained BART-large model when training each model within the QAG framework. We assume that comparative experiments using several sequence-to-sequence language models will be good future works.
Also, we only used six interrogative words, and did not consider `whose' and `whom' in the process. We considered these as originating from `who', but generating eight interrogative words including `whose' and `whom' would be a good approach. 
At last, in order to create a robust ranker, it is best to have a dataset that contains positive and negative samples. Since the manual data generation process required a time-consuming process, we utilize in-context negative samples as an alternative. If there is a dataset for the ranker learning purpose, much better performance can be achieved.

\section*{Ethics Statement}

\paragraph{Deployment} Our approach exploits parametric knowledge in the pre-trained model for language generation, which runs the risk of reflecting the bias of the training data. Undoubtedly, it is a well-known threat in tasks using a pre-trained model, but we must be more careful about social impact when using this method since our model aims to create educational QAs. Therefore, we plan to request model users to necessarily include a human review process of the generated QA pairs when used for educational purposes.

\paragraph{Human evaluation} We paid human workers more than the legal minimum wage. We also guided them to work remotely at any time they wanted and to rest when they are in a state of fatigue during work. Their B.A. degree certificate was discarded immediately upon confirmation to prevent personal information leakage. We made a task force to quickly respond to them if they have any questions or concerns by contacting them directly.

\section*{Acknowledgments}
We thank the anonymous reviewers for their valuable feedback and constructive suggestions. This work was supported by Hyundai Motor Company and Kia. This research was supported by the Ministry of Science and ICT (MSIT), Korea, under the Information Technology Research Center (ITRC) support program (IITP-2023-2018-0-01405) supervised by the Institute for Information \& Communications Technology Planning \& Evaluation (IITP) and
this work was supported by IITP grant funded by MSIT (No. 2020-0-00368, A Neural-Symbolic Model for Knowledge Acquisition and Inference Techniques) and this research was supported by MSIT, under the ICT Creative Consilience program(IITP-2023-2020-0-01819) supervised by the IITP.

\bibliography{custom}
\bibliographystyle{acl_natbib}

\appendix
\section{Overlap Mitigation Details}\label{app:overlap_algorithm}
The detailed process for overlap mitigation is as follows. We first define \textbf{Criterion} and \textbf{Metric}. \textbf{Criterion} means a sentence to be subject to overlap checking among questions or answers, and \textbf{Metric} means an evaluation metric to measure overlap. In our paper, we suggest using Rouge-L, or BLEU, as Metric.
In our main experiments, we choose criterion$_{i}$ as a$_{i}$ (\emph{i.e.} answer in QA pair), and Metric as ROUGE-L. In this process, Metric returns overlap score between 0 and 1. In estimating overlap, we lemmatize all the sentences and remove all the stop words in every QA pair.
\begin{algorithm}[h!]
    \caption{Overlapping based reranking}
    \label{alg:overlap}
    \begin{footnotesize}
        \begin{algorithmic}[1]
                \STATE \textbf{Given}: Passage $\textit{Psg}$, \\
                \STATE \textbf{Input}: Generated QA pair QA$^{gen}$ = $\{(q_{i}, a_{i})\}_{i=1}^{N}$ \\
                \STATE \textbf{Parameter}: int k \\
                \STATE \textbf{Define}: score$_{i}$ $\leftarrow$ RankingModule($q_{i}, a_{i}, \textit{Psg}$) \\
                \STATE \textbf{Choose}: criterion$_{i}$ $\leftarrow$ $q_{i}$ or $a_{i}$ \\
                \STATE \textbf{Choose}: Metric $\leftarrow$ ROUGE-L or BLEU \\
            \STATE output $\leftarrow$ [], comparing $\leftarrow$ []
            \WHILE{len(output) $\leq$ k}
                \FOR{ $(q_{j}, a_{j})$ in $QA^{gen}$}
                    \IF{comparing is not EMPTY}
                        \STATE overlaps$_{j}$ = [Metric(criterion$_{j}$, item) \\
                            \quad \quad \quad \quad \quad \quad \quad \quad \quad for item in comparing]
                        \STATE overlap$_{j}$ = max(overlaps$_{j}$)
                        \STATE \textbf{Define}: score$^{*}_{j}$ $\leftarrow$ score$_{j}$ - overlap$_{j}$ * |score$_{j}$|
                    \ELSE
                        \STATE \textbf{Define}: score$^{*}_{j}$ $\leftarrow$ score$_{j}$
                    \ENDIF
                \ENDFOR
                \STATE $(q_{i}, a_{i})$ $\leftarrow$ \textbf{Pick from} $QA^{gen}$ with highest score$^{*}_{j}$
                \STATE output $\leftarrow$ \textbf{Append} $(q_{i}, a_{i})$
                \STATE comparing $\leftarrow$ \textbf{Append} criterion$_{i}$
                \STATE $QA^{gen}$ $\leftarrow$ \textbf{Pop} $(q_{i}, a_{i})$
            \ENDWHILE
            \STATE \textbf{return} output
        \end{algorithmic}
    \end{footnotesize}
\end{algorithm}

\section{Implementation Performance on BART QGM and QAM}

\begin{table}[h]
\resizebox{0.5\textwidth}{!}{
\begin{tabular}{l|ccc|ccc}
\toprule[1.5pt]
& \multicolumn{3}{c|}{\textbf{QGM}}& \multicolumn{3}{c}{\textbf{QAM}} \\ \hline
\multicolumn{1}{c|}{} & \textbf{Rouge-L}& \multicolumn{1}{c}{\textbf{BLEU}} & \multicolumn{1}{c|}{\textbf{BERTScore}}& \textbf{Rouge-L}& \multicolumn{1}{c}{\textbf{BLEU}}  & \multicolumn{1}{c}{\textbf{BERTScore}}\\ \midrule[1.5pt]
\textbf{Ours}& \multicolumn{1}{c}{0.600} & \multicolumn{1}{c}{28.50} & \multicolumn{1}{c|}{0.934} & \multicolumn{1}{c}{0.542} & \multicolumn{1}{c}{43.29} & \multicolumn{1}{c}{0.936} \\ \bottomrule[1.5pt]
\end{tabular}
}
\caption{Performances on the BART QGM and QAM model.}\label{tb:QGQA_finetune}
\end{table}

In the iterative QAGen process, the QGM model and QAM model generate QA pairs, thereby obtaining a variety of QA pair candidates. We leverage FairytaleQA dataset for the model training, and results are shown in Table \ref{tb:QGQA_finetune}.

Our model utilizes a BART-large model identical to \citet{yao-etal-2022-ais}. Although the apples-to-apples comparison between the QGM model is impossible since ours are trained with interrogative word indicator, our QAM model performs slightly better than the result of \textsl{FQAG} (0.536 Rouge-L).

\section{Human Evaluation Details}\label{app:human_eval}
In our human evaluation process, all evaluators are degree holders in education or educational domain experts. We provide an evaluation sheet in the form of an API, and evaluators check the part corresponding to each question or write a rank order.

\begin{figure}[h]
\centering
\includegraphics[width=\linewidth]{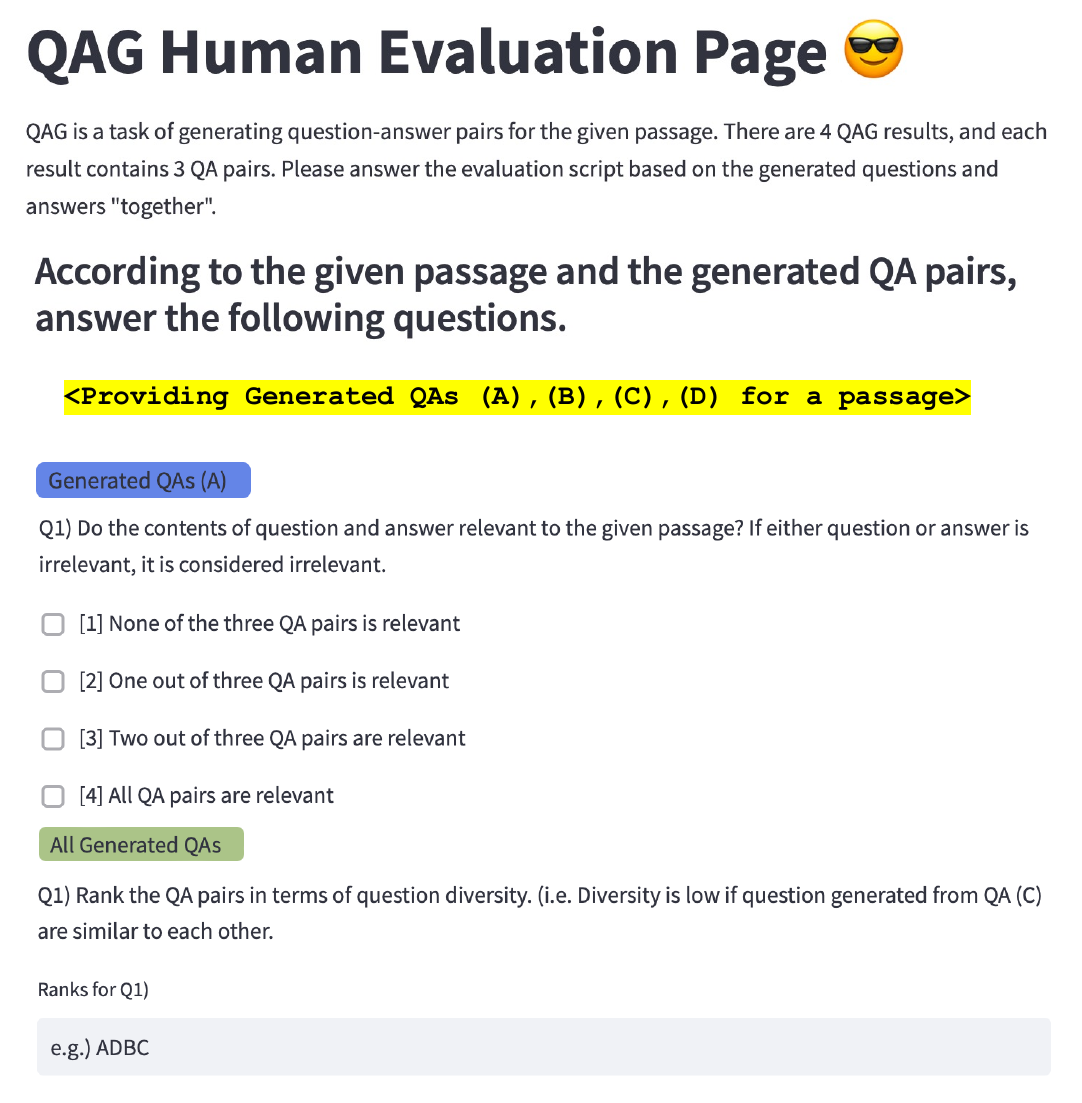 }
 \caption{Human evaluation script. We provide three QA pairs, each generated by the three systems and GT, and instruct the evaluators to score or rank them for a total of eight evaluation categories.}\label{fig:human_eval}
 \end{figure}
 
Figure \ref{fig:human_eval} describes the human evaluation script. In the local setting, we instruct evaluators to select how many of the three QA pairs produced by each system correspond to the given property. Rater responses are further converted to a 0-3 scale.
We also inform raters to rank four systems according to each of the three attributes per paragraph in the global setting. Both evaluations are performed for a total of 20 paragraphs in common. 
The intercoder-reliability scores (Krippendorff’s alpha)~\cite{krippendorff2011computing} among the evaluators are between 0.59 and 0.61. This is because each attribute of the criterion requires highly subjective assessment.

\section{QAGen Results}
To examine the practical applicability, we analyze the QA pairs generated by our framework. We compare our results to other QAgen systems, including \textsl{FQAG} and \textsl{SQG}, as well as the ground-truth QA pairs. We present the QAGen results for four passages, each containing three QA pairs. From a qualitative perspective, our framework generates QA pairs of higher quality compared to other systems.

\newpage

\begin{supertabular}{p{0.95\columnwidth}} 
\toprule[1.5pt] \\
\textbf{Passage \;} \\ 
Long, long ago japan was governed by hohodemi, the fourth mikoto (or augustness) in descent from the illustrious amaterasu, the sun goddess. He was not only as handsome as his ancestress was beautiful, but he was also very strong and brave, and was famous for being the greatest hunter in the land. Because of his matchless skill as a hunter, he was called "yama - sachi - hiko" or "the happy hunter of the mountains." \\ \midrule

\textbf{\textit{Ours}} \\
Q) What was yama-sachi-hiko called? \\
A) The happy hunter of the mountains.\\ \vspace{0.01cm} \\
Q) Why was he called "the happy hunter of the mountains"? \\
A) He was matchless in his skill as a hunter.\\ \vspace{0.01cm} \\
Q) What was special about hohodemi? \\
A) He was not only as handsome as his ancestress was beautiful, but he was also very strong and brave.\\ \midrule

\textit{\textbf{FQAG}} \\
Q) What was yama-sachi? \\
A) The happy hunter of the mountains.\\ \vspace{0.01cm} \\
Q) What was hohodemi called? \\
A) Yama-sachi-hiko.\\ \vspace{0.01cm} \\
Q) Who was the greatest hunter in japan? \\
A) The fourth mikoto.\\ \midrule

\textbf{\textit{SQG}} \\
Q) What was the name of the fourth mikoto? \\
A) Hohodemi \\ \vspace{0.01cm} \\
Q) Hohodemi was a descendant of what goddess? \\ 
A) Amaterasu \\ \vspace{0.01cm} \\
Q) Hohodemi was the fourth mikoto from what goddess? \\
A) Sun \\ \midrule

\textbf{\textit {GT}} \\
 
Q) Who governed japan long ago? \\
A) Hohodemi. \\ \vspace{0.01cm} \\
Q) What was special about hohodemi? \\
A) Handsome. \\ \vspace{0.01cm} \\
Q) Why was hohodemi called yama-sachi-hiko? \\ 
A) His matchless skill as a hunter. \\ \bottomrule[1.5pt]

\vspace{0.01cm} \\

\textbf{Passage \;} Then the dragon king interviewed the doctor and blamed him for not curing the queen. The doctor was alarmed at rin jin's evident displeasure, and excused his want of skill by saying that although he knew the right kind of medicine to give the invalid, it was impossible to find it in the sea. "Do you mean to tell me that you can't get the medicine here?" asked the dragon king. "It is just as you say!" said the doctor. "Tell me what it is you want for the queen?" demanded rin jin. "I want the liver of a live monkey!" answered the doctor. "The liver of a live monkey! Of course that will be most difficult to get," said the king. "If we could only get that for the queen, her majesty would soon recover," said the doctor. "Very well, that decides it; we must get it somehow or other. But where are we most likely to find a monkey?" asked the king. \\ \midrule

\textbf{\textit{Ours}}

Q) Where did the doctor say it was impossible to find the right kind of medicine to give the invalid? \\
A) In the sea. \\ \vspace{0.01cm} \\
Q) What happened after the dragon king interviewed the doctor and blamed him for not curing the queen? \\
A) The doctor was alarmed at rin jin's evident displeasure, and excused his want of skill by saying that although he knew the right kind \\ \vspace{0.01cm} \\
Q) Who did the doctor think would recover from the liver of a live monkey? \\
A) Her majesty. \\\midrule

\textbf{\textit{FQAG}}

Q) Who blamed the doctor for not curing the queen? \\
A) The dragon king. \\ \vspace{0.01cm} \\
Q) What did rin jinn want for the queen? \\
A) A live monkey. \\ \vspace{0.01cm} \\
Q) Who did the dragon king ask where they would find a monkey's liver? \\
A) The king. \\\midrule

\textbf{\textit{SQG}} 
Q) Who interviewed the doctor and blamed him for not curing the queen? \\
A) Dragon king \\ \vspace{0.01cm} \\
Q) Who blamed rin jin for not curing the queen? \\
A) The dragon king \\ \vspace{0.01cm} \\
Q) Who did the dragon king blame for not curing the queen? \\
A) Rin jin \\\midrule

\textbf{\textit{GT}}
Q) Who was blamed for not curing the queen? \\
A) The doctor. \\ \vspace{0.01cm} \\
Q) Why was the liver of a live monkey difficult to get? \\
A) They are under water. \\ \vspace{0.01cm} \\
Q) How did the doctor feel when he was blamed by the king? \\
A) Alarmed. \\ \bottomrule[1.5pt]

\vspace{0.01cm} \\

\textbf{Passage \;} \\
After a time they all came to a town where a king reigned whose daughter was so serious and solemn that no one could ever manage to make her laugh. So the king had decreed that whoever should succeed in making her laugh should marry her. When dullhead heard this he marched before the princess with his goose and its appendages, and as soon as she saw these seven people continually running after each other she burst out laughing, and could not stop herself. Then dullhead claimed her as his bride, but the king, who did not much fancy him as a son-in-law, made all sorts of objections, and told him he must first find a man who could drink up a whole cellarful of wine. Dullhead bethought him of the little grey man, who could, he felt sure, help him; so he went off to the forest, and on the very spot where he had cut down the tree he saw a man sitting with a most dismal expression of face. \\\midrule

\textbf{\textit{Ours}} \\
Q) Where did dullhead see the man sitting with a dismal expression of face? \\
A) The very spot where he cut down the tree. \\ \vspace{0.01cm} \\
Q) What happened when the princess saw the seven people continually running after each other? \\
A) She burst out laughing and could not stop herself. \\ \vspace{0.01cm} \\
Q) Who did the king want dullhead to find before he could marry the princess? \\
A) A man who could drink up a whole cellarful of wine. \\\midrule

\textbf{\textit{FQAG}} \\
Q) Who reigned in a town where no one could ever manage to make the daughter laugh? \\
A) A king. \\ \vspace{0.01cm} \\
Q) Who could help dullhead?
A) The little grey man. \\ \vspace{0.01cm} \\
Q) Where did dullhead go to find the little grey man? \\
A) The forest. \\\midrule

\textbf{\textit{SQG}} \\
Q) What did the king decree that whoever succeeded in making her laugh should do? \\
A) Marry her \\ \vspace{0.01cm} \\
Q) How many people were running after each other? \\
A) Seven \\ \vspace{0.01cm} \\
Q) Where did dullhead go to find a man who could help him? \\
A) The forest \\\midrule

\textbf{\textit{GT}} \\
Q) Who did the king decree should marry his daughter? \\
A) Whoever should succeed in making her laugh. \\ \vspace{0.01cm} \\
Q) How will the little grey man help dullhead? \\
A) Drink up a whole cellarful of wine. \\ \vspace{0.01cm} \\
Q) How did the king feel about dullhead as a son-in-law? \\
A) Unhappy. \\ \bottomrule[1.5pt]

\vspace{0.01cm} \\

\textbf{Passage \;} Many, many years ago there lived a good old man who had a wen like a tennis-ball growing out of his right cheek. This lump was a great disfigurement to the old man, and so annoyed him that for many years he spent all his time and money in trying to get rid of it. He tried everything he could think of. He consulted many doctors far and near, and took all kinds of medicines both internally and externally. But it was all of no use. The lump only grew bigger and bigger till it was nearly as big as his face, and in despair he gave up all hopes of ever losing it, and resigned himself to the thought of having to carry the lump on his face all his life. \\\midrule

\textbf{\textit{Ours}} \\
Q) What did the good old man have? \\
A) A wen like a tennis-ball growing out of his right cheek. \\ \vspace{0.01cm} \\
Q) How long did the old man have the wen like a tennis-ball growing out of his right cheek? \\
A) Many, many years. \\ \vspace{0.01cm} \\
Q) Where did the lump grow out of? \\
A) His right cheek. \\\midrule

\textbf{\textit{FQAG}} \\
Q) Who had a wen like atennis-ball growing out of his right cheek? \\
A) The old man. \\ \vspace{0.01cm} \\
Q) Where did the lump grow? \\
A) His right cheek. \\ \vspace{0.01cm} \\
Q) What did the old man do to get rid of his lump? \\
A) He consulted many doctors far and near. \\\midrule

\textbf{\textit{SQG}} \\
Q) What type of ball did the old man have a wen like? \\
A) Tennis \\ \vspace{0.01cm} \\
Q) What was the wen like a tennis - ball growing out of his right cheek to the old man? \\
A) Great disfigurement \\ \vspace{0.01cm} \\
Q) What did the old man try to get rid of the lump? \\
A) Everything \\\midrule

\textbf{\textit{GT}} \\
Q) Why was the man not able to get rid of his wen? \\
A) The doctors did not know how to get rid of it. \\ \vspace{0.01cm} \\
Q) How did the man feel about his wen? \\
A) Annoyed. \\ \vspace{0.01cm} \\
Q) What did the good old man have growing in his right cheek? \\
A) A wen. \\ \bottomrule[1.5pt] \\

\hspace{0.5cm} \\
\end{supertabular}

\appendix


\end{document}